**APPLIED RESEARCH**

# Characterization of Magnetic Labyrinthine Structures Through Junctions and Terminals Detection Using Template Matching and CNN

VINÍCIUS YU OKUBO[1], KOTARO SHIMIZU[2], B. S. SHIVARAM[3], AND HAE YONG KIM[1]
[1]Department of Electronic Systems Engineering, Polytechnic School, University of São Paulo, São Paulo 05508-010, Brazil
[2]Center for Emergent Matter Science, RIKEN, Saitama 351-0198, Japan
[3]Department of Physics, University of Virginia, Charlottesville, VA 22904, USA

Corresponding author: Vinícius Yu Okubo (ViniciusOkubo@usp.br)

This work was supported in part by the Coordenação de Aperfeiçoamento de Pessoal de Nível Superior-Brasil (CAPES)-Finance Code 001. The work of B. S. Shivaram was supported in part by NSF at the University of Virginia under Grant DMR #2016909.

**ABSTRACT** Defects influence diverse properties of materials, shaping their structural, mechanical, and electronic characteristics. Among a variety of materials exhibiting unique defects, magnets exhibit diverse nano- to micro-scale defects and have been intensively studied in materials science. Specifically, defects in magnetic labyrinthine patterns, called junctions and terminals are ubiquitous and serve as points of interest. While detecting and characterizing such defects is crucial for understanding magnets, systematically investigating large-scale images containing over a thousand closely packed junctions and terminals remains a formidable challenge. This study introduces a new technique called TM-CNN (Template Matching - Convolutional Neural Network) designed to detect a multitude of small objects in images, such as the defects in magnetic labyrinthine patterns. TM-CNN was used to identify 641,649 such structures in 444 experimental images, and the results were explored to deepen understanding of magnetic materials. It employs a two-stage detection approach combining template matching, used in initial detection, with a convolutional neural network, used to eliminate incorrect identifications. To train a CNN classifier, it is necessary to annotate a large number of training images. This difficulty prevents the use of CNN in many practical applications. TM-CNN significantly reduces the manual workload for creating training images by automatically making most of the annotations and leaving only a small number of corrections to human reviewers. In testing, TM-CNN achieved an impressive F1 score of 0.991, far outperforming traditional template matching and CNN-based object detection algorithms.

**INDEX TERMS** Computer vision, convolutional neural networks, deep learning, magnetic labyrinthine patterns, material science, object detection, template matching.

## I. INTRODUCTION

Most materials exhibit some kind of order which is typically characterized by a certain order parameter maintaining coherence or constancy over long distances compared to the atomic scale. The degrees of freedom within the order parameter directly influences the array of properties manifested by the material's phases. Specifically in magnetic materials, phases can be characterized by the alignment of the magnetic moments. These magnetic moments can assume a variety of orders, such as ferromagnetics, where the magnetic moments are uniformly aligned and antiferromagnetics,

The associate editor coordinating the review of this manuscript and approving it for publication was Roberto C. Ambrosio.

where the moments alternate site by site [1]. Apart from orders directly proportional to lattice spacings, incommensurable orders are also frequently observed and documented [2], [3], [4], [5], [6], [7]. One illustrative example is the stripe phase, characterized by spatial modulation in the orientations of magnetic moments, with a period distinct from an integer multiple of the lattice spacing.

The above magnetic phases do not necessarily appear in a coherent manner and are often accompanied by defects. As an example illustrating the breakdown of spatial coherence, we present magnetic domain images of bismuth doped yttrium iron garnet (Bi:YIG) films at zero field in Fig. 1, with intensity reflecting the out-of-plane component of the magnetic moments. In this material, although the most stable order is the uniform stripe phase, the magnetic domains show non-periodic labyrinthine structures resulting from the propagation of stripes in various directions accompanied by a plethora of defects seemingly present everywhere [8], [9], [10], [11], [12]. However, these complex labyrinthine patterns do possess discernible characteristics which demand proper quantification.

We present in Fig. 1 two images both obtained in zero applied magnetic field. In Fig. 1a, which we label as the "quenched" state, the borders of dark and bright domains exhibit a sinuous nature and do not appear as parallel. In contrast, the "annealed" state, shown in Fig. 1b, consists of regions with nearly parallel domains. This state exhibits roughly equal widths of dark and bright domains, and the areas occupied by them are also approximately equal for any sampled region. Therefore, the stripes in the annealed state show greater spatial coherence.

Within these magnetic structures, defects take the form of interruptions in the stripes known as "terminals" and points where multiple stripes conjoin, referred to as "junctions" (Fig. 2). Since defects directly affect the spatial coherence of the ordered phase, the number and correlations of defects can serve as a crucial metric for quantifying the deviation of a structure from a perfectly periodic structure, namely the stripe order. Furthermore, such defects have also been gathering considerable attention due to their implications on physical phenomena arising from the nontrivial geometric properties [13], [14], [15]. Thus, in the realm of condensed matter physics, experimental identification of the number and positions of such defects plays an important role in characterizing material properties.

Accurately counting and differentiating genuine structures from misidentifications is crucial to a quantitative physical understanding of the origins and evolution of these patterns. Manual annotation of defects is infeasible. For instance, we used 444 images with 641,649 structures [16]. Furthermore, manual annotation relies on subjective interpretation of junctions and terminals, which could lead to counting inconsistencies. In order to address these issues, automated processes are required. Algorithms for finding objects in an image are known as object detectors and can be broadly

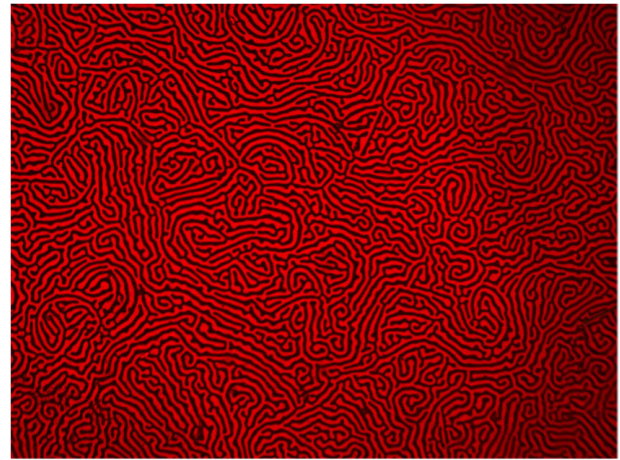

(a) Quenched state.

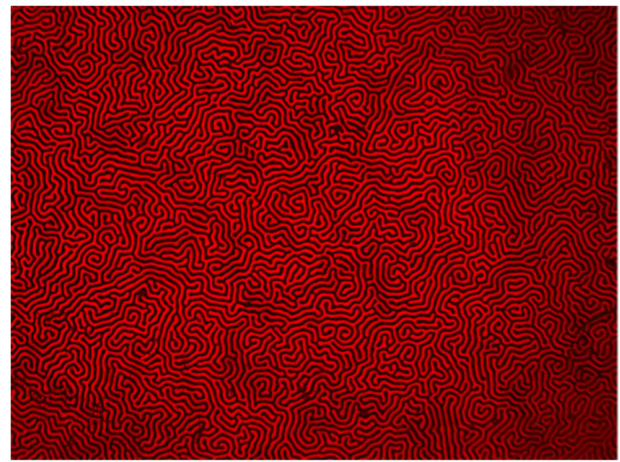

(b) Annealed state.

**FIGURE 1.** Examples of magnetic labyrinthine patterns.

categorized into classical methods and those based on deep learning.

### A. CLASSICAL OBJECT DETECTION METHODS
Classical object detection methods span a variety of techniques to extract and process features from the image. For instance, template matching [17] is a technique employed to find a pre-defined template within a larger image. This is achieved by scanning the entire image and calculating the correlation between the template and the scanned region. Viola and Jones object detecting algorithm [18] employs multiple template matchings using Haar-like features, each of them serving as a weak detector. These features are ensembled through a boosting strategy to form a strong detector. The Histogram of Oriented Gradients (HOG) [19] represents another approach to extract useful features, dividing the image into cells and calculating a gradient histogram for each. This extracted information can be used by machine learning algorithms, such as support vector machine, to identify detections.

## B. DEEP LEARNING OBJECT DETECTION METHODS

In the last decade, deep learning approaches have revolutionized the field of computer vision, surpassing traditional machine learning techniques in multiple image processing tasks [20], [21]. For object detection, Girshick et al. introduced R-CNN [22], marking it as one of the pioneering detection techniques rooted in deep learning. R-CNN operates as a classification-based model: multiple regions are extracted from an image and each is classified independently. This straightforward method led to significant improvements in detection, achieving new state-of-the-art results on the Pascal VOC dataset [23], compared to earlier techniques like Haar features and HOG. A distinguishing feature of R-CNN is its region proposal step. Directly processing every conceivable region of varying sizes and positions in an image is computationally impractical. Hence, this step selects a simplified set of regions from the original image for individualized classification by the CNN model. Further developments brought by Faster R-CNN [24] have improved both accuracy and speed by integrating the region proposal into the model.

Redmon et al. introduced YOLO [25], a deep-learning detection approach modeled as a regression task. This method partitions the image into a grid and each cell contains their own set of outputs. The grid cell where the object is centered has the task of identifying the position, dimensions and class of the object. A standout benefit of this approach is its efficiency: YOLO processes the image in a single pass, contrasting with R-CNN-based models that are divided into region proposal and classification steps. This significantly reduces inference time, enabling real-time video detection. However, YOLO was not able to achieve the precision and recall rates of Faster R-CNN when tested on the Pascal VOC dataset. Comparisons in small object detection settings have also shown Faster R-CNN to outperform YOLO [26].

Further architectural developments have also improved detection. Mask R-CNN by He et al. [27] extends the capabilities of Faster R-CNN by simultaneously performing object detection and instance segmentation. Cascade R-CNN by Cai and Vasconcelos [28] proposes a multi-stage architecture with sequentially higher thresholds to refine detection quality. RetinaNet by Lin et al. [29] introduces Focal Loss to address the imbalances between foreground and background classes coupled with dense detection and a Feature Pyramid Network architecture for improving detection at different scales [30].

To address the limitations of anchor-based architectures, such as their sensitivity to hyper-parameter selection, anchor-free architectures were proposed. Law and Deng proposed CornerNet [31], framing detections as paired corner key-points. FCOS by Tian et al. [32] applies detection at the feature level by applying a correspondence between feature coordinates and separate areas from the image. The feature coordinates where their corresponding area intersects with an object are tasked for detection. Sparse R-CNN by Sun et al. [33] is inspired by Faster R-CNN, but uses learned detection proposals instead of predicted ones.

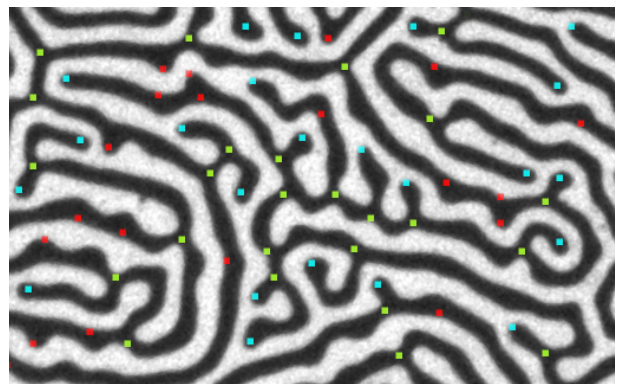

**FIGURE 2.** Defects in the magnetic labyrinth patterns. Junctions are marked in green and terminals in cyan. Red dots indicate false positives from the template matching, which are filtered out by the CNN classifier.

Recently, the transformer architecture, originally introduced for language processing tasks, has been adapted for use in computer vision [34], [35]. Carion et al. [36] developed DETR, utilizing transformers to perform object detection without the need for custom-designed components. Alternatively, Liu et al. [37] introduced Swin, a variation of the transformer for feature extraction which outperforms CNN-based approaches in various vision tasks, demonstrating the potential of transformers to enhance object detection capabilities. DINO, proposed by Zhang et al. [38] is a variant of DETR which uses deformable attention [39] and adds noise to the ground truth during training, enabling the architecture to surpass traditional multi-stage detectors.

## C. THE PROBLEM AND PROPOSED SOLUTION

Modern digital microscopy allows the easy acquisition of high-resolution experimental images, which contain intricate regular patterns and a large number of defects, sometimes numbering in the thousands. This makes it difficult to use deep learning, as creating an accurately annotated dataset would be a laborious process due to the large number of defects. Furthermore, common deep learning models such as YOLO and Faster R-CNN were not designed to detect thousands of small objects. Benchmarks with both methods show that performance degrades when detecting small objects, with Faster R-CNN maintaining a slight advantage over YOLO [26].

Correlation-based granulometry is a technique that can be used to detect a large number of small objects in the image. It was proposed by Maruta et al. [40], [41] to analyze the distribution of square and circular pores in the macroporous silicon layer in scanning electron microscope images. This technique is called ''granulometry'' because the objective of the original application was to obtain a histogram of pore distribution as a function of size. It was later used by Araújo et al. [42] to detect individual bean grains, analyze each grain, and calculate the quality of the bean batch. This technique performs multiple template matchings to achieve robustness against angle and shape variations, and then performs non-maximum suppression to avoid finding the same object multiple times. However, this technique could not be applied directly in our problem, because defects in labyrinthine magnetic structures vary greatly in shape and cannot be accurately detected using only template matchings.

To address these challenges, we developed a new two step technique that we named Template Matching - Convolutional Neural Network (TM-CNN), which integrates correlation-based granulometry with the CNN classifier. First, a series of template matchings detects potential candidates for junctions and terminals. By setting a low threshold, we cast a broad net for candidates detection, minimizing false negatives but increasing false positives. These candidates are filtered by non-maximum suppression to avoid multiple detections of the same defect. Second, a CNN classifier filters out the false positives.

We demonstrate that this method substantially outperforms template matching alone, while streamlining the image annotation process and reducing the computational burden typically associated with deep learning detection techniques. We also show that the TM-CNN technique outperforms Faster R-CNN, achieving a significantly higher F1-score in junctions and terminals detection. TM-CNN executable files for Windows are available for download for demonstration purposes.[1]

### D. STRUCTURE OF THE ARTICLE

The remainder of this work is organized as follows. Section II presents existing techniques for detecting junctions and terminals, differentiating classical and deep learning approaches. Section III describes the dataset used and the proposed TM-CNN technique in detail. Section IV presents experimental results from the detection of defects in magnetic labyrinthine patterns and discusses their significance in understanding physical phenomena. Finally, Section V concludes the article by reflecting on the merits of our work.

## II. RELATED WORKS
### A. JUNCTIONS AND TERMINALS DETECTION

Junctions and terminals are shapes with relevance that extend beyond materials. Within computer vision, recognizing and enumerating them has been performed in diverse contexts, such as natural landscapes [43], biology [44] and handwriting images [45].

### B. CLASSICAL METHODS FOR JUNCTIONS AND TERMINALS DETECTION

One prevalent approach for detecting junctions and terminals involves using skeletonization as a pre-processing step. This process reduces the image to one-pixel-wide lines to represent its structures. Points in skeleton can then be identified as terminals, junctions and crossings based on their neighboring pixels. This technique has been applied in vascular images [46] and in the analysis of handwritten Chinese characters [47].

[1] https://github.com/okami361/TM-CNN/releases/tag/Windows_v2.3.7

Pre-processing techniques using contour information have also been explored for junction detection. Lee and Wu [48] investigated stroke extraction in Chinese characters. Their method segments regions according to their contour, identifying junctions by counting neighboring regions. Maire et al. [49] proposed a junction detector in natural images by locating intersecting contours. This approach applies an expectation–maximization style algorithm to iteratively select relevant contours and suggest the junction's position.

However, skeletonization and contour finding are noise-sensitive processes and a pre-processing error will lead to a detection error.

Junctions and terminals can also be identified by analyzing the arrangement of linear structures within the image. Su et al. [50] describe a technique for identifying these linear structures using the Hessian Matrix. This approach has been validated in biological images such as blood vessels, neutrites and tree branches. Xia et al. [43] present a junction detection method in natural images, based on amplitudes and phases of the normalized gradients of the image.

Template-based approaches quantify the similarity of the appearance of image regions and the template. Deriche and Blaszka [51] modeled this approach as energy minimization, which is calculated by the deviation between the image and a predetermined model. This enabled the detection of key image features, such as edges, corners and terminals.

### C. DEEP LEARNING METHODS FOR JUNCTIONS AND TERMINALS DETECTION

Owing to the success of R-CNN based detection techniques, Pratt et al. [44] developed a pipeline for identifying junctions and crossings in retinal vascular structures. Their method involves two main steps: initially, detection regions are proposed centred along the blood vessels, which are then classified as junctions, crossings or background. To generate the detection regions, their approach requires a binary segmented version of the exam. These images undergo a skeletonization process, with the resulting points serving as references for the centers of the blood vessels.

Zhao et al. [52], addressing the same problem, proposed using a Mask R-CNN based model [27] for region proposal. This strategy enables inference without the need for binary segmented version of the exam. However, during training, the segmented images are still used in the Mask R-CNN model to enhance its learning capabilities. This approach surpassed the performance of previous techniques in the detection of junctions and crossings in retinal vascular images.

### D. DEFECTS DETECTION IN MAGNETIC LABYRINTHINE IMAGES

In the context of magnetic labyrinthine patterns, a previous work employed a notably smaller dataset, consisting only of several dozen defects, for manual defect detection [10].

Recently, persistent homology has been used to extract the topological features of the labyrinthine patterns in a systematic way [53]. In this method, a persistent diagram is constructed for binarized magnetic domain images to extract the geometric characteristics of their spatial structure. By combining the persistent diagram with principal component analysis, one can map the real-space distribution of junctions, terminals, and bending points within the labyrinthine patterns, while their classification remains difficult.

In this work, we achieved the systematic detection and classification, enabling us to investigate the labyrinthine patterns in a quantitative manner. To characterize the evolution of the labyrinthine patterns, we applied the TM-CNN technique to 444 images and investigated how the number of defects and their locations change step by step.

## III. METHODOLOGY
### A. DATASET

In this study, we used films of a ferromagnetic material of recognized technological importance, bismuth doped yttrium iron garnet, or Bi:YIG, and obtained magnetic images of the labyrinthine patterns using a microscope with polarized light [54]. We specifically focused on the evolution of the labyrinthine patterns under the demagnetization field protocol described below. First, we prepared the sample in the fully saturated state by applying a sufficiently large magnetic field in $+z$ direction, which is perpendicular to the films. In this state, magnetic moments in the Bi:YIG film are forced to point in the field direction. Next, we instantaneously dropped the field to zero and hold it for 10 seconds to get the image. Since the magnetic field is zero, magnetic moments can point upward and downward, resulting in the labyrinthine patterns shown in Fig. 1; the bright and dark regions represent the domains with opposite directions of magnetic moments. Such a process involving switching on and off the magnetic fields is considered half of the demagnetization step. In the remaining half step, a magnetic field was applied with a reduced amplitude and oriented in the opposite direction. We again captured the magnetic domain image after reducing the magnetic field to zero. By repeating these protocols up to 18 steps, we investigated the evolution of the labyrinthine patterns in the demagnetization process step by step. The amplitude of the magnetic field was exponentially reduced with each step. We conducted a series of demagnetization processes from the fully saturated state, repeating this cycle 6 times. Furthermore, we explored another situation, where the magnetic field was initially applied in the $-z$ direction, and its direction was alternated step by step. Consequently, a total of 12 demagnetization processes were performed, yielding a collection of 444 domain images. All measurements reported here were performed at room temperature. The experimentally obtained images covered an area of 2 mm × 1.8 mm.

The original high-resolution color images were converted to grayscale and their resolutions were reduced to 1300 × 972 to facilitate processing. Furthermore, a median filter with kernel size 3 was applied to reduce noise.

### B. TM-CNN OVERVIEW

Our approach to detect junctions and terminals in magnetic labyrinthine patterns consists of two sequential steps: proposal of potential detections and their classification between junction, terminal and false detections. It is inspired by other cascaded object detection techniques like Viola and Jones face detection [18], R-CNN [22], and scale and rotation invariant template matching [55].

Fig. 3 illustrates the overall structure of TM-CNN. In the first phase, the algorithm generates a preliminary set of potential detections. It must propose all true defects, even if it also generates many false positives. We achieve this by applying template matching detection with a low threshold, followed by a non-maximum suppression. In the second phase, in order to eliminate the false positives, each potential detection is filtered by a CNN classifier.

### C. TEMPLATE MATCHING

The basic form of template matching finds instances of a smaller template $T$ within a larger image $I$. This is done by calculating some similarity metric between the model $T$ and the content of a moving window located at each possible position of $I$. We measured the similarity between the image and the template using the Normalized Cross Correlation (NCC). NCC is invariant to linear changes in brightness and/or contrast. NCC between template $T$ and image $I$ at pixel $(x, y)$ is calculated as:

$$NCC_{T,I}(x,y) = \frac{\sum_{x',y'}(\tilde{T}(x',y') \cdot \tilde{I}(x+x', y+y'))}{\sqrt{\sum_{x',y'}\tilde{T}(x',y')^2 \cdot \sum_{x',y'}\tilde{I}(x+x', y+y')^2}}$$
(1)

where $(x', y')$ is the coordinate inside the template, $\tilde{T}(x', y')$ is the mean-corrected template value at $(x', y')$, and $\tilde{I}(x, y)$ is the value of the mean-corrected image inside the moving window at pixel $(x, y)$. Mean-correction consists of subtracting the average value from each pixel. Template matching using NCC can be implemented very efficiently using Fast Fourier Transform (FFT) [56].

This basic approach is not well suited for detecting junctions and terminals in the magnetic labyrinthine structures because a single template is not capable of modeling:

1) All possible rotations;
2) All deformed shapes of defects.

To solve problem, we employ a rotation-invariant template matching based on exhaustive evaluation of rotated templates. There are some alternative rotation-invariant techniques based on circular and radial projections [55], [57], and on Fourier coefficients of circular and radial projections [58]. These techniques can reduce computational requirements, but their implementations are complex and require parameter tuning. Furthermore, our application does not require exceptional computational performance, as the processing is offline. Thus, our technique uses the standard OpenCV[2] template matching implementation, which is highly optimized using FFT and special processor instructions.

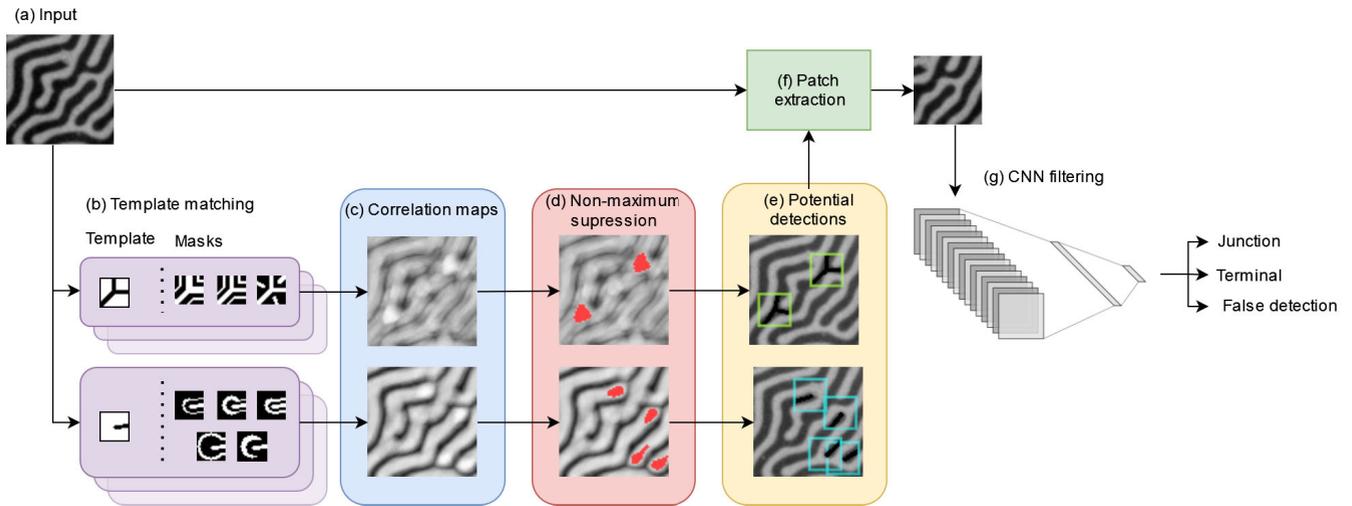

**FIGURE 3.** TM-CNN overview: (a) The algorithm receives a magnetic labyrinthine image. (b) It applies template matchings to compute correlation maps. (c) These maps measure the similarity between the templates and each region of the image. (d) TM-CNN looks for high values in the maps and applies non-maximum suppression. (e) The potential detections are located at the highest correlation values. (f) The algorithm extracts patches centered on each detection. (g) A CNN classifies the detection as junction, terminal or false detection.

Problem arises from complex shape variations of defects in magnetic labyrinthine patterns. Typically, terminals and junctions do not align perfectly with the templates, mainly because real magnetic strips have non-uniform curvatures and widths. Template matching allows for slight shape variations. However, to further improve flexibility, we use masks. Many template matching implementations allow you to specify a mask, besides the template, indicating which pixels in the template will not be considered in the correlation calculation. White pixels in the mask indicate significant template pixels, and black pixels indicate template pixels to disregard ("don't care"). If the mean-corrected template value at some pixel $(x', y')$ is zero, that is $\tilde{T}(x', y') = 0$, this pixel will not be taken into account to compute NCC. On the other hand, one cannot use FFT to speed up the NCC calculation if it contains conditional executions. Therefore, it is very likely that optimized implementations will assign zero to the template pixels corresponding to the black mask pixels, after the mean correction.

### D. TEMPLATES AND MASKS USED IN THE EXPERIMENT
We manually designed templates and masks, and empirically tuned them to capture various junction and terminal shapes. They have a resolution of 21 × 21 and are generated at runtime. Templates represent magnetic strips as black lines radiating from their center drawn on a white background. Meanwhile, masks are created from black backgrounds with white areas defining relevant coordinates used in template matching. Their main purpose is to obscure the space between strips and background, addressing variations in widths and curvatures, and to limit the background regions to reduce interference from neighboring strips. Table 1 exemplifies the templates and masks used to detect junctions and terminals.

[2]Open Source Computer Vision Library, https://opencv.org.

**TABLE 1.** Templates and masks used to detect junctions and terminals.

| | Template | Mask | Mask description |
|---|---|---|---|
| Junction | | | |
| Terminal | | | |
| | Effective strip | Effective background | Discarded region |

We created 439 junction templates to accommodate variations in shapes and rotations. Junction templates consist of three radial lines at rotation $\alpha, \beta, \gamma \in \mathbb{N}$ with $0° \leq \alpha < \beta < \gamma < 360°$ (Table 1). These templates are drawn iteratively in steps of 15°, while maintaining a difference of at least 70° and at most 190° between two consecutive angles. For each triple of parameters $(\alpha, \beta, \gamma)$, a single template is generated along with three masks: two masks "type A" where the effective background is parallel to strips and one mask "type B" where the effective background is positioned between the strips (Table 2). Masks are created according to the rotations $\alpha, \beta, \gamma$ and width values $w_{bg}, w_{bs}, w_{sp} \in \mathbb{N}$, representing the width of the effective background, the width between effective background and strip, and the width of the effective strip, respectively (Table 1).

TABLE 2. Collection of masks used in our study (measurements in pixels).

| | $w_{bg}$ | $w_{bs}$ | $w_{sp}$ | $R_{bg}$ | $R_{bs}$ | Mask |
|---|---|---|---|---|---|---|
| Junction ($\alpha=60°, \beta=180°, \gamma=300°$) | | | | | | |
| junction_mask 1 (type A) | 1 | 7 | 3 | - | - | |
| junction_mask 2 (type A) | 3 | 5 | 2 | - | - | |
| junction_mask 3 (type B) | -6 | 6 | 2 | - | - | |
| Terminal ($\alpha=30°$) | | | | | | |
| terminal_mask 1 | 2 | 3 | 2 | 5 | 2 | |
| terminal_mask 2 | 1 | 5 | 2 | 5 | 3 | |
| terminal_mask 3 | 0 | 4 | 2 | 7 | 4 | |
| terminal_mask 4 | 1 | 6 | 2 | 10 | 8 | |
| terminal_mask 5 | 0 | 7 | 2 | 8 | 4 | |

Terminal templates are simpler and use a single rotation parameter $\alpha \in \mathbb{N}$, varying within $0° \leq \alpha < 360°$ in steps of 3°, totaling 120 variants. To model variations in terminal and background shapes, we created five masks for each model. Terminal masks are determined by six parameters, including the previous four ($\alpha, w_{bg}, w_{bs}, w_{sp}$), and the two additional radius parameters $R_{bs}, R_{bg}$ that define the two circles at the tip of the terminal, reflecting the strip and background respectively.

Template matching is applied separately for each (template, mask) pair. Several template matchings are computed in parallel using the OpenMP library.[3] This process takes about 80 seconds to process an image on an i7-9750H processor.

To obtain the final correlation map $corr$, we calculate the maximum value among all $n = 3 \times 439 + 5 \times 120 = 1917$ NCC maps for each position $(x, y)$, that is:

$$corr(x, y) = \text{MAX}_{i=1}^{n}[\text{NCC}_i(x, y)] \quad (2)$$

Pixels where $corr(x, y)$ exceed a predefined threshold $t$ are considered potential detections. However, a single junction/terminal may encompass multiple neighboring points with correlation values greater than the threshold $t$. Therefore, it is necessary to perform some form of non-maximum suppression to eliminate duplicate detections and select only the true center of junction/terminal. Kim et al. [41] present a solution to this problem: Whenever two potential detection points $p_1$ and $p_2$ are separated by a distance smaller than a threshold $c$, the point with the lowest correlation value is discarded. In this work, we use a slightly different approach, but with the same practical result: Whenever the algorithm finds a potential detection, it executes a breadth-first search algorithm. This algorithm recursively searches adjacent pixels $(x, y)$ where the correlation exceeds 80% of the threshold (that is, $corr(x, y) > 0.8\,t$) and saves the pixel with the highest correlation. Subsequently, the searched area has its correlation value set to zero to avoid re-detection. Fig. 3d highlights the searched area in red. The pixel with the highest correlation is chosen as the center of the junction/terminal. This process performs detection in a single pass.

---
[3]Open Multi-Processing, https://www.openmp.org/.

### E. DATASET ANNOTATION
To classify potential detections into true or false using a CNN classifier, we must first create annotated training images. TM-CNN makes it easier to create training examples, as it allows one to annotate the examples semi-automatically. This process is divided in two phases.

#### 1) TEMPLATE MATCHING-ASSISTED ANNOTATION
In this phase, only a small set of images are annotated. Initially, we apply template matching followed by non-maximum suppression to identify the centers of possible detections, together with their probable labels (junction or terminal). Without this help from template matching, we would have to manually and precisely locate the centers of thousands of defects. Subsequently, a human reviewer makes corrections to ensure that the labels given by the template matching are correct. The reviewer may change the labels to junction, terminal or false detection. After all positive detections are annotated along with a small set of false detections, a larger set of false detections is created by lowering the template matching threshold and sampling new false detections. These images, now with positive and negative annotations, are used to train a preliminary version of the CNN classifier.

#### 2) DEEP LEARNING-ASSISTED ANNOTATION
Due to the small number of images in the initial training set, the preliminary CNN classifier cannot accurately classify all magnetic stripe defects. Nonetheless, this preliminary model is integrated into the annotation procedure to alleviate the required workload. In the second phase, we continue using template matching to generate the initial set of detections. However, the preliminary CNN classifier is employed to identify most of the template matching errors, thus speeding up the annotation process. As new images are annotated, more accurate models are trained to further simplify the annotation workload. The final training set consists of 17 images derived from a single annealing protocol, selected to cover varied experimental configurations of ascending and descending magnetic fields at different magnitudes. Out of these, 16 were selected from the quenched (unordered) state, as they represent a more diverse set of shapes and represent a greater challenge for classification. The training set encompasses a total of 33,772 detections, which includes 12,144 junctions, 12,777 terminals, and 8,851 false detections.

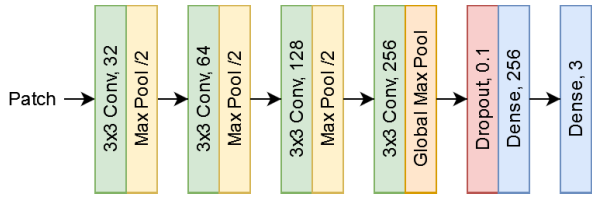

FIGURE 4. CNN architecture to classify a patch in junction, termination or false detection.

### F. CANDIDATE FILTERING BY CNN

Our algorithm extracts small $50 \times 50$ patches centered around each detection point and a CNN classifies them into three classes: junction, terminal or false positive. The size of patches for CNN classification is larger than the size of template matching models ($21 \times 21$), allowing CNN to use more contextual information than template matching.

We use a simple CNN model to classify the small patches (Fig. 4). It has four convolutional layers with 32, 64, 128 and 256 filters, all using $3 \times 3$ kernels. The first three convolutional layers are followed by max pooling layers to downsample the feature maps and global max pooling is applied after the last convolutional layer. This is followed by dropout and two fully connected layers: the first with 128 nodes and the second with three output nodes. The ReLU activation function is used across the model, except at the output layer where the softmax is used. In total, this network has only 422,608 parameters. For comparison, VGG-16 and ResNet-50 (common backbones for detection) have 138 million and 25 million parameters, respectively. Thanks to its simplicity, our model can make predictions even without GPUs and takes around 30 seconds for filtering each image using an i7-9750H processor and 16 GB of RAM with no dedicated graphics or AI accelerators.

We implemented this model in Python using the TensorFlow library.[4] To enhance rotational invariance, we employed data augmentation techniques, performing random rotations of the images in multiples of 90°. We pre-trained it using the MNIST [59] dataset for 15 epochs. We then fine-tuned the model for our application for another 20 epochs. We used categorical cross-entropy as the loss function, and the model was optimized using Adam with the learning rate of $10^{-3}$. The model was trained using the Google Colab environment[5] on a single V100 GPU.

## IV. EXPERIMENTS AND RESULTS
### A. PERFORMANCE EVALUATION

Evaluating TM-CNN poses a certain challenge because there are no benchmarks or published results for direct comparison. So, we defined that junction and terminal detections reviewed by human observers as the gold standard and compared the algorithms against this standard. For testing purposes, we annotated 10 images from a separate experimental trial, using the same procedure as the training images (subsection III-E). False detections were excluded since they do not contribute to the detection evaluation. In total, the test set contained 7,374 junctions and 7,431 terminals.

[4]https://www.tensorflow.org/
[5]https://colab.google/

TABLE 3. Performance of different algorithms to detect junctions and terminals.

| Model | Precision | Recall | F1-Score |
|---|---|---|---|
| TM only | $0.825 \pm 0.102$ | $0.934 \pm 0.182$ | $0.876 \pm 0.145$ |
| **TM-CNN** | $\mathbf{0.993 \pm 0.003}$ | $\mathbf{0.989 \pm 0.004}$ | $\mathbf{0.991 \pm 0.002}$ |
| Faster R-CNN | $0.462 \pm 0.019$ | $0.327 \pm 0.015$ | $0.383 \pm 0.017$ |
| $CLS_{Conv4}$ [60] | $0.768 \pm 0.013$ | $0.650 \pm 0.020$ | $0.704 \pm 0.013$ |

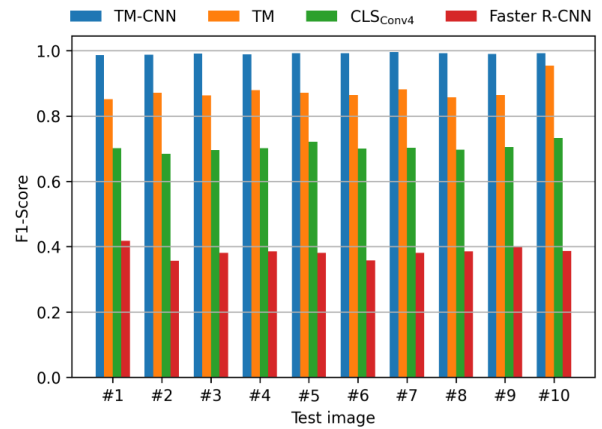

FIGURE 5. F1-score obtained for each test image.

We tested four algorithms: template matching only (our TM-CNN algorithm, without CNN from the second stage); the proposed TM-CNN algorithm; the widely used Faster R-CNN; and its variant $CLS_{Conv4}$ proposed by Eggert et al. that improves small object detection [60]. The Faster R-CNN model had to be adapted, since it was originally designed to detect large objects. We used a single anchor of size $24 \times 24$, as all defects are approximately this size. The model's backbone was VGG16, in line with Ren et al. [24]. This backbone was initially pre-trained on the MNIST dataset for digit classification. We fine-tuned the model for junction and terminal detection for 25 epochs using the 17 training images, with learning rate of $10^{-5}$ and the Adam optimizer.

We used precision, recall and F1-score as evaluation metrics. A detection was considered positive if it achieved Intersection over Union (IoU) greater than 0.5, and the detection threshold was adjusted to optimize the F1-score. Table 3 presents the averages and standard deviations of the performance metrics across the 10 tests, showing that TM-CNN substantially outperforms template matching only, Faster R-CNN and its variant $CLS_{Conv4}$. To demonstrate the consistency of the results, Fig. 5 displays the F1-scores obtained for each test image by the four algorithms. TM-CNN always present F1-score superior to the other three algorithms. TM-CNN achieves detection accuracy close to 100% and most mistakes are in detections where a human observer would also be in doubt about the correct classification (Fig. 6).

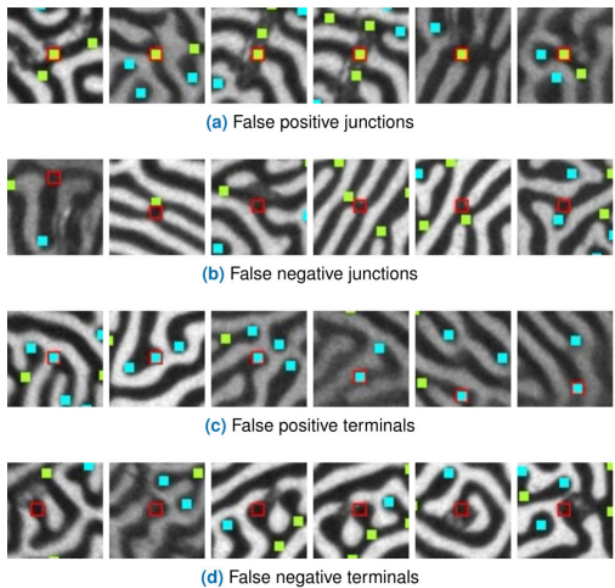

**FIGURE 6.** Examples of TM-CNN detection errors, highlighted in red. Green and cyan dots are respectively junctions and terminals detected by TM-CNN.

### B. RESULTS FROM A PHYSICS PERSPECTIVE

In Bi:YIG films, it is known that a perfect stripe pattern is the most energetically favorable and stable configuration of magnetic moments [10]. In this structure, the magnetic moments exhibit stripes aligning straight in one direction throughout the entire structure, resulting in large spatial coherence. Meanwhile, the experimentally obtained structures often manifest labyrinthine patterns, as illustrated by dark and bright stripes in Fig. 1. In these labyrinthine patterns, stripes are aligned within small domains but propagate in different directions between different domains, and thus the spatial coherence of stripes is smaller than the perfect stripe. Notably, junctions and terminals are defects emerging in the formation of these labyrinthine patterns, intricately linked to the growth of spatial coherence in the system. Introducing such defects into the perfect stripe structure leads to the bending and branching of the initially aligned stripes, eventually reducing spatial coherence accompanied by the transformation to labyrinthine patterns. Indeed, a plethora of defects are observed in the labyrinthine patterns shown in Fig. 2. Conversely, the elimination of these defects plays a crucial role in the transition from a quenched state to an annealed state, where the system increases the spatial coherence. Hence, the evolution of the labyrinthine patterns would be systematically quantified through the number of junctions and terminals.

These defects fall within a distinctive category known as topological defects. Each defect cannot be eliminated alone by continuous deformation of the stripes due to their topological properties, while they can be removed in pairs. For example, in Fig. 2, terminals branch from junctions and can be removed in pairs, reducing the branching distance. The property that defects cannot be removed and created individually by continuous deformation arises from geometric constraints and has become one of the important features of topological defects [13], [15]. Given that the stripe order emerges from labyrinthine patterns by removing defect, one might intuitively anticipate that the number of junctions and terminals would be comparable and decrease with increasing the spatial coherence.

Fig. 7 illustrates the step dependence of the number of junctions and terminals during the demagnetization process. In the experiment, two types of processes were performed: one where the magnetic field starts from the upwards direction, referred to as the positive process, and the other where it starts from the downwards direction, referred to as the negative process. The results for each are shown in Figs. 7a and 7b, respectively.[6]

Results from six different experimental runs were averaged for each process to estimate the number of defects. In the initial three steps, the numbers of junctions and terminals increase with steps from $\sim$ 750 to $\sim$ 850. After that, the numbers sharply decrease to $\sim$ 700 around step 5 and remain nearly unchanged from step 10 onward. Consequently, the transition to the quenched state is inferred to start before step 5 and be completed by step 10. The reduction in the number of defects during the demagnetization process is compatible with the naive expectation based on the physical argument that the number of defects would decrease in association with the increase in the spatial coherence of the stripes. Furthermore, in each process, the numbers of junctions and terminals are within the margin of error bars for both processes. This is consistent with the topological argument that junctions and terminals are paired. We note that these numbers are comparable between the positive and negative processes, which serves as collateral evidence that our TM-CNN algorithm appropriately detects defects.

Now, we delve into the evolution of the number of junctions and terminals from the energy landscape perspective. In our experiments, diverse metastable labyrinthine patterns were observed, yet the energetically most stable stripe patterns were never observed. The transitions between these different metastable states can be analogously understood as traversing potential hills to move from one valley to another in the complex energy landscape of Bi:YIG. Realizing the stripe order corresponds to finding the deepest valley within it. In the initial stage of the demagnetization process, a large magnetic field is applied to "shake" the system, thus facilitating transitions between different valleys in pursuit of lower energy states. Consequently, the number of defects changes significantly in the first ten steps, where the transition from the quenched state to the annealed state takes place. However, progressing into the latter stage, the number of defects remains unchanged as the amplitude of the field decreases. This result indicates the presence of numerous metastable states, each of which is deep, making transitions less likely for smaller fields. Further quantification of the evolution of the labyrinthine patterns based on the spatial distribution of the defects (and not only on their numbers) is discussed in another article [16].

---

[6]As described in Sec. III-A, imaging is conducted upon quenching the magnetic field to zero. Within a single step of the positive process, the magnetic field is first applied upwards, followed by quenching, and then applied downwards before another quenching. The field direction is reversed for the negative process. Therefore, two images are obtained at each step. In Fig. 7, we present the number of defects in the images obtained after quenching from the negative field for the fair comparison between the positive and negative processes.

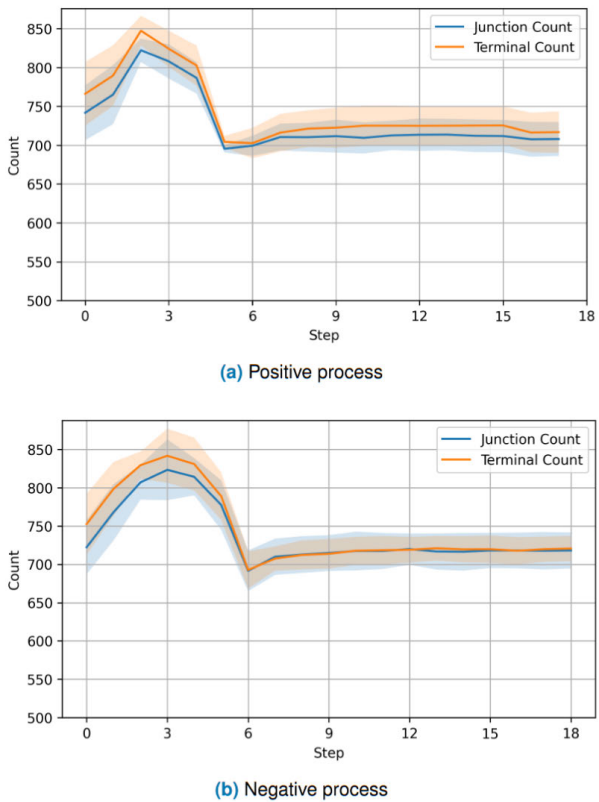

**FIGURE 7.** Experimental averages of junction and terminal counts per step with one standard deviation shaded for the demagnetization processes starting from (a) the positive and (b) negative magnetic fields (see the texts for details).

## V. CONCLUSION

In this work, we presented a new algorithm named TM-CNN to detect defects in magnetic labyrinthine patterns, contributing to a pioneering analysis in material science. Our study characterized the evolution of junctions and terminals in magnetic stripes during demagnetization procedures, aiming at better understanding defect arrangement in magnetic materials [16].

TM-CNN employs a two-stage detection procedure, combining template matching for initial detection and a convolutional network classifier for refining misdetections. This approach ensures a high detection accuracy and facilitates dataset annotation through a semi-automatic procedure.

In our experiments, TM-CNN exhibited performance superior to other techniques, achieving an impressive F1 score of 0.991. This high performance is mainly due to TM-CNN's ability to locate small and clustered objects. TM-CNN achieves almost 100% accuracy with a simple CNN classifier with less than half a million parameters and can be used even on computers without GPUs.

However, since TM-CNN uses manually defined templates and masks for object representation, it may not be ideal for detecting objects with complex shape variations. In general, the varieties of defects in materials depend on the degrees of freedom of the order parameters, giving rise to diverse defects with distinct morphologies. Consequently, additional enhancements are necessary to achieve impartial detection of diverse defects. Additionally, our technique is much slower than most modern object detectors, many of them designed for real-time object detection. Addressing these challenges could further enhance TM-CNN's applicability.

While TM-CNN was developed for defect detection in labyrinthine magnetic patterns, its potential applications are not limited to this field. Future research could explore the use of TM-CNN in other domains, such as identifying bifurcations in blood vessels or adapting it to other structures that can be modeled using templates. In addition, correlation analysis based on these detections would be an important future task, specifically from the viewpoint of Physics. Taking junctions and terminals as examples, the distance between defects along the domains is known to characterize transformations of labyrinthine patterns [11]. Such a correlation analysis is also eagerly awaited to characterize the material properties in a more accurate and statistical manner.

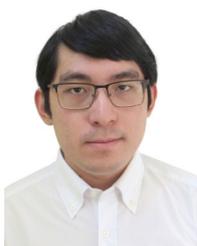

**VINÍCIUS YU OKUBO** was born in Campinas, Brazil, in 1999. He received the B.S. degree in electrical engineering from the University of São Paulo (USP), in 2022, where he is currently pursuing the M.S. degree in electrical engineering. His research interests include computer vision and deep learning.

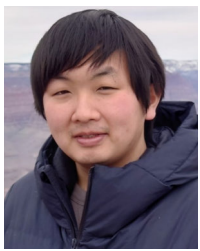

**KOTARO SHIMIZU** was born in Tokyo, Japan, in 1996. He received the B.S. degree in physics from Waseda University, Japan, in 2019, and the M.S. and Ph.D. degrees in physics from The University of Tokyo, Japan, in 2021 and 2024, respectively. He is currently a Theoretical Physicist working in condensed matter physics. His main research interests include magnetism, strongly correlated electrons, and topological phases of matter.

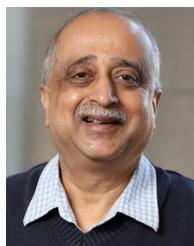

**B. S. SHIVARAM** received the B.S. degree in physics, chemistry and mathematics from Bangalore University, India, in 1977, the M.S. degree in physics from the Indian Institute of Technology, Madras, India, in 1979, and the Ph.D. degree in experimental condensed matter physics from Northwestern University, Evanston, IL, USA, in 1984. From 1985 to 1986, he was a Research Associate with the James Frank Institute, University of Chicago, USA. Since 1987, he has been a Faculty Member with the Department of Physics, University of Virginia, Charlottesville, VA, USA. He is currently an Experimental Physicist working in a broad area of condensed matter including magnetism. He was a recipient of the Alfred P. Sloan Foundation Fellowship. He serves as the Vice-Chair for the Forum for Outreach and Engaging the Public of the American Physical Society.

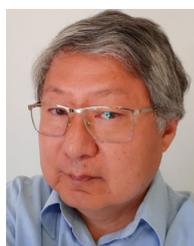

**HAE YONG KIM** was born in South Korea. He received the B.S. and M.S. degrees (Hons.) in computer science and the Ph.D. degree in electrical engineering from the Universidade de São Paulo (USP), Brazil, in 1988, 1992, and 1997, respectively.

He is currently an Associate Professor with the Department of Electronic Systems Engineering, USP. He is the author of more than 100 articles and holds three patents. His research interests include image processing, machine learning, medical image processing, and computer security.

Dr. Kim and colleagues received the 6th edition of the Petrobras Technology Award in the "Refining and Petrochemical Technology" Category, in 2013; the Best Paper in Image Analysis Award from the Pacific-Rim Symposium on Image and Video Technology, in 2007; and the Thomson ISI Essential Science Indicators "Hot Paper" Award, for writing one of the top 0.1% of the most cited computer science papers, in 2005.

• • •